\documentclass[11pt]{article}

\usepackage[final]{acl}

\usepackage{latexsym}
\usepackage[utf8]{inputenc}
\usepackage{inconsolata}
\usepackage{times}
\usepackage{booktabs,siunitx}
\usepackage{minitoc}
\usepackage{microtype}      
\usepackage{xcolor}         
\usepackage{xspace}
\usepackage{graphicx}
\usepackage{amsmath}
\usepackage{amssymb}
\usepackage{multicol}
\usepackage{float}
\usepackage{array}
\usepackage{framed}
\usepackage{mdframed}
\usepackage{tikz} 
\usepackage{multicol,multirow}
\usepackage{setspace}
\usepackage{makecell}
\usepackage{xcolor}
\usepackage{caption}

\usepackage{subcaption} 
\usepackage[ruled,vlined]{algorithm2e}

\usepackage[most,skins,theorems]{tcolorbox}

\usepackage{enumitem}     
\usepackage{bm} 
\definecolor{paleblue}{HTML}{F0F8FF}

\usepackage{colortbl}
\usepackage{arydshln}
\usepackage[textsize=tiny]{todonotes}
\definecolor{newgreen}{rgb}{0.7, 0.9, 0.7}
\definecolor{newblue}{rgb}{0.85, 0.85, 0.9}
\usepackage{amssymb} 
\usepackage{cleveref}

\newcommand{\bskblue}[1]{\textcolor{black}{#1}}
\newcommand{\bskorange}[1]{\textcolor{black}{#1}}
\usepackage[table]{xcolor}
\definecolor{tkcolor}{RGB}{224,223,255}
\definecolor{lgcolor}{RGB}{240,240,240}

\definecolor{darkgreen}{rgb}{0.0, 0.5, 0.0} 
\definecolor{lightblue}{RGB}{220,235,250}
\definecolor{lightgray}{RGB}{211,211,211}
\definecolor{lightgreen}{RGB}{230, 255, 230}
\definecolor{mygray}{gray}{0.85}  

\newcommand{\cred}[1]{{$_{#1}$}}
\newcommand{\cgreen}[1]{{$_{#1}$}}

\title{TTVS: Boosting Self-Exploring Reinforcement Learning via Test-time Variational Synthesis}

\author{Sikai Bai, Haoxi Li, Jie Zhang\footnotemark[2], Yongjiang Liu, Song Guo\footnotemark[2] \\
  The Hong Kong University of Science and Technology \\
  \texttt{\{sbaiae, hligb, yliunr\}@connect.ust.hk},\\ \texttt{\{csejzhang, songguo\}@cse.ust.hk} \\}

\begin{document}
\maketitle
\renewcommand{\thefootnote}{\fnsymbol{footnote}} 
\footnotetext[2]{Corresponding authors.}
\begin{abstract} \label{01abstract}
Despite significant advances in Large Reasoning Models (LRMs) driven by reinforcement learning with verifiable rewards (RLVR), this paradigm is fundamentally limited in specialized or novel domains where such supervision is prohibitively expensive or unavailable, posing a key challenge for test-time adaptation. 
While existing test-time methods offer a potential solution, they are constrained by learning from static query sets, risking overfitting to textual patterns. 
To address this gap, we introduce Test-Time Variational Synthesis (TTVS), a novel framework that enables LRMs to self-evolve by dynamically augmenting the training stream from unlabeled test queries. TTVS comprises two synergistic modules: (1) Online Variational Synthesis, which transforms static test queries into a dynamic stream of diverse, semantically-equivalent variations, enforcing the model to learn underlying problem logic rather than superficial patterns; (2) Test-time Hybrid Exploration, which balances accuracy-driven exploitation with consistency-driven exploration across synthetic variants. Extensive experiments show TTVS yields superior performance across eight model architectures. Notably, using only unlabeled test-time data, TTVS not only surpasses other test-time adaptation methods but also outperforms state-of-the-art supervised RL-based techniques trained on vast, high-quality labeled data.

\end{abstract}
\section{Introduction} \label{02Introduction}

Recent advances in Large Reasoning Models (LRMs) have demonstrated remarkable capabilities, achieving significant breakthroughs on complex reasoning tasks such as mathematics~\citep{yang2024qwen2,guo2025deepseek} and programming~\citep{hui2024qwen2,khan2025llm}. A primary driver behind this progress is the paradigm of Reinforcement Learning with Verifiable Rewards (RLVR), which has been proven by frontier models like OpenAI-o1~\citep{jaech2024openai} and DeepSeek-R1~\citep{guo2025deepseek} to endow LLMs with advanced reasoning capabilities. A key factor in these developments is the availability of large-scale supervised datasets~\citep{bai2025qwen2,dubey2024llama}, which incorporate instructions and corresponding ground-truth labels for reward computation.

However, in specialized domains such as clinical diagnostics~\citep{ullah2024challenges,mcduff2025towards} and aerospace engineering~\citep{liu2025llm,connolly2025development}, collecting high-quality data and verifiable supervision remains a significant challenge. Acquiring such human-annotated data is often prohibitively expensive and difficult to scale~\citep{villalobos2024position}. While a potential alternative is to use external LLMs for cost-effective data generation, this approach usually depends on access to highly capable expert-level models, which are not always readily available~\citep{yang2023gpt4tools,goldie2504synthetic}. As complex and unlabeled problems continuously emerge, this reliance on external supervision becomes a fundamental bottleneck, motivating a transition to what~\citet{silver2025welcome} call the ``era of experience'', where models are enabled to self-evolve without direct human oversight.

Building upon this vision, it naturally motivates a promising direction in which LRMs autonomously improve via RL on unlabeled data. In this paper, we focus on a particularly potent mode of such self-evolution: adaptation to test-time data~\citep{shu2022ttt,snell2024tts,zhang2025ttssurvey}. While recent methods have successfully enabled test-time RL by generating rewards through majority voting~\citep{zuo2025ttrl}, 
they are fundamentally constrained by learning from a fixed, predefined set of test queries. Namely, this approach still operates on a static test set, which may lead to overfitting on superficial textual patterns rather than the underlying problem logic. This raises a pivotal question: \textit{if reward signal can be generated on-the-fly, can the training data itself be dynamically augmented to enhance performance?}


To answer this question, we propose \textbf{Test-Time Variational Synthesis (TTVS)}, a novel framework that boosts self-exploring RL by actively transforming static test queries into a dynamic training stream. Specifically, TTVS is composed of two synergistic modules: (1) The \textit{Online Variational Synthesis} module leverages the model's own consensus to generate semantically-equivalent but differently phrased queries, applying an online filter to ensure quality and diversity. This process enforces the model to learn the underlying problem logic rather than superficial patterns. (2) The \textit{Test-time Hybrid Exploration} module utilizes a dual-mode update strategy, Intra-Group Exploration (IGE) for accuracy-driven exploitation and Cross-Group Exploration (CGE) for consistency-driven exploration, to enable robust and generalizable learning within this augmented data space.
Moreover, TTVS is agnostic to policy optimization algorithms and can be flexibly incorporated into other methods, such as GRPO~\cite{shao2024grpo}, OPO~\cite{hao2025opo}, and DAPO~\cite{yu2025dapo}.
Extensive experiments across various benchmarks and model families (Qwen3, Qwen2.5, LLaMA) demonstrate that TTVS achieves superior performance in mathematical reasoning using solely unlabeled test-time data. Remarkably, TTVS not only surpasses other test-time counterparts but also outperforms state-of-the-art RL-based post-training methods, where the latter depend on large-scale, annotated datasets.

\section{Related Works}\label{03relatedworks}
\textbf{Reinforcement Learning (RL)} \cite{sutton1998rl}
is a pivotal technique for advancing the reasoning capabilities of Large Language Models (LLMs). Early research focused on Reinforcement Learning from Human Feedback (RLHF) \cite{christiano2017RLHF, ouyang2022RLHF}, which aligns LLMs with human values by training a reward model based on human preference data. Despite its efficacy, RLHF is often resource-intensive, necessitating extensive human annotation \cite{gao2023issueRLHF}. 
Recently, Reinforcement Learning with Verifiable Rewards (RLVR) \cite{schulman2017ppo, luong2024RLVR} has emerged as a more scalable alternative for domains with objective correctness criteria (e.g., mathematics reasoning) using rule-based reward functions instead of humans. 
Despite the progress shown by state-of-the-art RL algorithms (e.g., GRPO \cite{shao2024grpo}, DAPO \cite{yu2025dapo}), a fundamental challenge persists: these models are trained via supervision, but their inference during test-time necessitates CoT reasoning for novel and unseen problems.

\noindent\textbf{Test-time Scaling (TTS)} \cite{zhang2025ttssurvey} aims to improve model performance by leveraging additional computational resources at inference time. It is primarily divided into two paradigms: \textit{1) Test-Time Inference (TTI)} \cite{wang2022tti, yuan2024tti} enhances output quality by generating multiple solution candidates and selecting the best one via a scoring function or iterative refinement. \textit{1) Test-Time Training (TTT)} \cite{osowiechi2024ttt, zhang2025ttt} adapts a pre-trained model to distribution shifts at inference time, commonly employing self-supervised objectives such as entropy minimization and pseudo-labeling. However, prior work has primarily focused on video generation \cite{dalal2025video} and understanding \cite{liu2024video}. Existing TTS work on the integration of LLMs and reinforcement learning remains limited \cite{fang2025self}, and prevailing methods rely on offline training with fixed test sets, which prevents the model from achieving continual self-improvement that adapts to its current capabilities.

\noindent\textbf{Self-improvement  Reinforcement Learning} 
aims to overcome the cost and label requirements of traditional Reinforcement Learning (RL) \cite{tian2024selfimprovement}. Current approaches largely fall into two categories: \textit{1) Synthetic data generation} \cite{goldie2025synthetic,kim2025synthetic} utilizes powerful expert models (e.g., DeepSeek R1 and GPT-4)  to generate synthetic data. While effective, this approach incurs substantial computational costs and relies on external models. \textit{2) Self-rewarding metric} \cite{bai2022selfreward,zhang2025selfreward, zuo2025ttrl} uses predefined rules to generate preference pairs for RL training or adopts a majority-vote-based self-rewarding/certainty mechanism in an unsupervised RL setting. However, a critical flaw of the self-rewarding strategy is its reliance on the fixed dataset. 

\section{Preliminaries}
\label{sec:preliminaries}

\subsection{Reinforcement Learning for LLMs}
We frame the task of improving LLM reasoning as a reinforcement learning problem. An LLM is treated as a policy $\pi_\theta$, parameterized by $\theta$, which generates a sequence of tokens (a solution) $o = (t_1, t_2, \dots, t_L)$ in response to an input query $q$. The goal of RL is to optimize the parameters $\theta$ to maximize the expected reward for the generated solutions:
\begin{equation}
    \label{eq:rl_objective}
    \max_{\theta} \mathbb{E}_{q \sim \mathcal{D}, o \sim \pi_\theta(\cdot|q)} [R(o, q)],
\end{equation}
where $\mathcal{D}$ is the data distribution and $R(o, q)$ is a reward function that evaluates the quality of the solution $o$ for the query $q$.

\subsection{Group Relative Policy Optimization}
In this work, we adopt Group Relative Policy Optimization (GRPO)~\cite{shao2024grpo}, an efficient RL algorithm well-suited for LLM training as it forgoes an explicit critic model. For a given query $q$, GRPO samples a group of $G$ outputs $\{o_1, \dots, o_G\}$ from the current policy $\pi_\theta$. The policy is then updated by maximizing the following objective:
\begin{equation}
    \label{eq:grpo}
    \mathcal{J}_{\text{GRPO}}(\theta) = \mathbb{E} \left[ \frac{1}{G}\sum_{i=1}^{G} \mathcal{L}_i(\theta) \right],
\end{equation}
where $\mathcal{L}_i(\theta)$ is the per-sample objective, often a clipped surrogate objective similar to PPO, weighted by the advantage $A_i$:
\begin{small}
\begin{equation}
    \label{eq:grpo_loss}
    \mathcal{L}_i(\theta) = \min \left( \rho_i(\theta), \text{clip}(\rho_i(\theta), 1-\epsilon, 1+\epsilon)) A_i \right).
\end{equation}
\end{small}

Here, $\rho_i(\theta) = \pi_\theta(o_i|q)/\pi_{\theta_{\text{old}}}(o_i|q)$ is the probability ratio. A key feature of GRPO is that it estimates the advantage $A_i$ directly from the group's rewards $\{r_1, \dots, r_G\}$ by treating the mean reward as a baseline, which is both computationally efficient and effective:
\begin{equation}
    \label{eq:grpo_advantage}
    A_i = \frac{r_i - \text{mean}(\{r_j\}_{j=1}^G)}{\text{std}(\{r_j\}_{j=1}^G) + \delta},
\end{equation}
where $\delta$ is a small constant for numerical stability.

\subsection{Reward Signal via Majority Voting}
\label{sec:majority_vote}
In the test-time setting, ground-truth labels are unavailable, making reward computation a central challenge. Following TTRL~\citep{zuo2025ttrl}, we generate a reward signal by estimating a pseudo-label via majority voting. Given a query $q$, we first sample $N$ candidate solutions $\{o_1, \dots, o_N\}$ from the policy. We then extract a final answer $a_i = \textrm{ExtractAnswer}(o_i)$ from each solution. The most frequently occurring answer (\textrm{MajorityVote}$(\cdot)$) is designated as the pseudo-label $y^*$:
\begin{equation}
    \label{eq:maj_vote}
    y^* = \underset{a}{\text{argmax}} \sum_{i=1}^N \mathbb{I}(a_i = a),
\end{equation}
where $\mathbb{I}(\cdot)$ is the indicator function. The reward $r_i$ for each solution $o_i$ is then determined by its agreement with this pseudo-label:
\begin{equation}
    \label{eq:reward_func}
    r_i = R(o_i, y^*) = 
    \begin{cases} 
        1, & \text{if } a_i = y^* \\
        0, & \text{otherwise.}
    \end{cases}
\end{equation}

This mechanism provides a robust, self-generated reward signal that enables RL training in the absence of explicit supervision.
\section{Methodology} \label{method}
\begin{figure*}[!t]
    \center
    \includegraphics[width=0.93\textwidth]{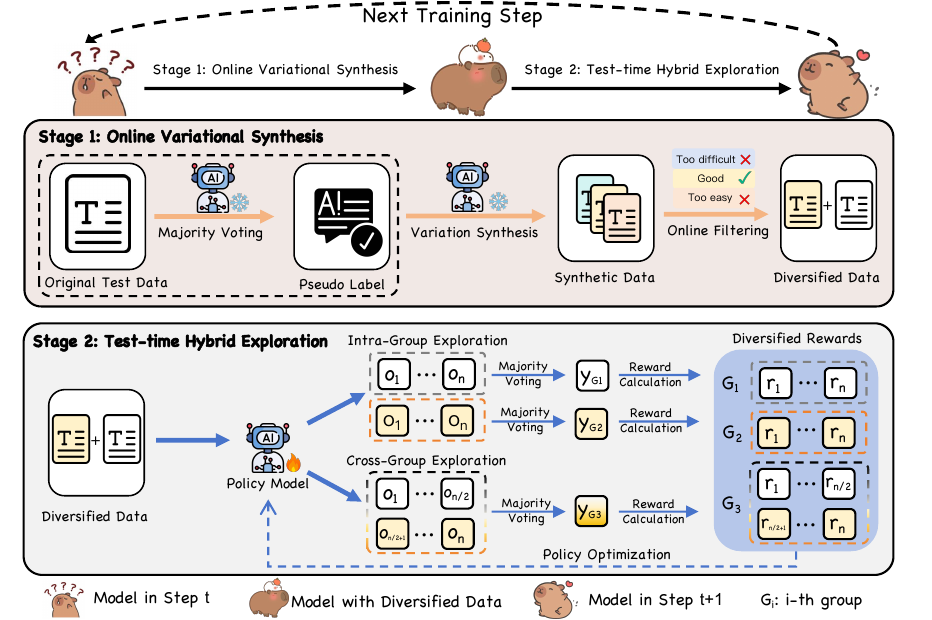}
    \caption{
    The schematic illustration of the Test-time Variational Synthesis (TTVS).
    In \textbf{Stage 1: Online Variational Synthesis.} Pseudo labels are generated from test data to synthesize diversified samples, which are then online filtered for quality and diversity. In \textbf{Stage 2: Test-time Hybrid Exploration.} A policy model is optimized on this augmented data through a hybrid intra- and cross-group exploration strategy, culminating in diversified rewards that guide the model's optimization.
    }
    \label{fig:intro}
    \vspace{-0.2cm}
\end{figure*}

In this work, we propose TTVS, a novel test-time RL framework that actively augments the training data and employs a dual-update mechanism to balance accuracy and generalization. Our method enables the model not only to adapt to the test data but also to learn the invariant underlying logic of the problems. The overall pipeline is depicted in Figure~\ref{fig:intro} and consists of two primary stages: (1) Online Variational Synthesis and (2) Test-time Hybrid Exploration.

\subsection{Online Variational Synthesis}
The first stage of our framework focuses on generating a rich and diversified set of training instances from the original, static test query. This is achieved through a three-step process: pseudo-label generation, variational data augmentation, and online filtering.

We first generate a pseudo-label for each original query $q$ from the test set. Given a query $q$, we sample $N$ candidate solutions (rollouts) $\{o_1, o_2, \dots, o_N\}$ from the current policy $\pi_\theta$. An answer $a_i$ is extracted from each rollout, and a consensus answer $y_q^*$ is determined via majority voting. This consensus answer serves as the pseudo-label for the query $q$.
\begin{equation}
    \{o_1, \dots, o_N\} \sim \pi_\theta(\cdot|q),
\end{equation}
\begin{equation}
    y_q^* = \textrm{MajorityVote}((a_i)_{i=1}^N),
\end{equation}
where pseudo-label $y_q^*$ is the cornerstone for both reward calculation and the subsequent data synthesis process.

At the heart of our method lies the concept of online variational synthesis. Instead of solely relying on the original query $q$, we prompt the policy model $\pi_\theta$ to generate $k$ new queries that are semantically consistent with $q$ and share the same answer $y_q^*$, but differ in their surface-level expression. We design a specific prompt $\mathcal{P}$ that instructs the model to paraphrase the original problem. The generation process is as follows:
\begin{equation}
    \{q'_1, q'_2, \dots, q'_k\} \sim \pi_\theta(\cdot | \mathcal{P}, q, \ y_q^*).
\end{equation}

This synthesis process transforms a single data point into a cluster of semantically equivalent problems $\{q, q'_1, \dots, q'_k\}$. Training on this augmented data encourages the model to develop a more robust and abstract understanding of the reasoning task, moving beyond mere pattern matching of the input text.


To ensure the quality of the augmented data, we introduce a crucial online filtering stage. Not all synthesized queries are retained. A query cluster originating from $q$ is only generated and used for training if it satisfies two conditions. First, the initial group accuracy for the original query $q$, denoted $\text{acc}(q)$, must lie within a predefined difficulty range $[\tau_{\text{low}}, \tau_{\text{high}}]$:
\begin{equation}
    \label{eq:filter_acc}
    \text{acc}(q) = \frac{1}{N} \sum_{i=1}^{N} \mathbb{I}(a_i = y_q^*) \in [\tau_{\text{low}}, \tau_{\text{high}}].
\end{equation}

This condition ensures that the model focuses on problems that are neither trivial nor intractable. Second, each synthesized query $q'_j$ must not exceed a maximum token length $L_{\text{max}}$:
\begin{equation}
    \label{eq:filter_len}
    \texttt{Length}(q'_j) \le L_{\text{max}},
\end{equation}
only the data satisfying these criteria proceed to the policy update stage, thus curating a training batch of suitable difficulty and quality. We provide more details for online filtering in Appendix \ref{online_filtering}

\subsection{Test-time Hybrid Exploration}
Having generated a high-quality batch of original and variational queries $\mathcal{D}_{\text{batch}}$, the second stage performs policy updates using our proposed hybrid exploration mechanism.  Two complementary update modes, Intra-Group Exploration and Cross-Group Exploration, are designed to execute independently within each training step. The rationale behind this dual-mode approach is to synergistically improve both the model's accuracy on specific problem instances and its consistency across semantic variations.

The first mode, Intra-Group Exploration (\textbf{IGE}), focuses on maximizing performance on each individual problem by operating within the confines of a single query's generated outputs. For every query $\tilde{q} \in \mathcal{D}_{\text{batch}}$ (where $\tilde{q}$ can be an original query $q$ or a variational query $q'_j$), we perform a full, independent GRPO update. The process begins by performing a complete rollout to generate $N$ solutions, from which a pseudo-label $y_{\tilde{q}}^*$ is obtained via intra-group majority voting. Subsequently, the advantage for each rollout is calculated based on this intra-group reward signal $R(o_i, y_{\tilde{q}}^*)$, and the policy parameters $\theta$ are updated by optimizing the GRPO objective for that specific query $\tilde{q}$.

This mode represents an accuracy-driven exploitation of the data, as it treats each query as a distinct optimization target to be mastered. It pushes the model to find the correct solution for that specific phrasing. The total update from this mode can be seen as the sum of independent policy gradients for all queries in the batch.

The second mode, Cross-Group Exploration (\textbf{CGE}), is designed to explicitly enforce consistency by operating across the semantically-equivalent problem cluster $\{q, q'_1, \dots, q'_k\}$. Instead of performing separate rollouts for each query, we create a mixed pool of rollouts by sampling a fraction of solutions from each query in the cluster:
\begin{equation}
    \mathcal{O}_{\text{mix}} = \bigcup_{j=0}^{k} \left\{ \text{Sample}_{i=1}^{N/(k+1)} \left( o_i \sim \pi_\theta(\cdot|q'_j) \right) \right\},
\end{equation}
where $q'_0$ denotes the original query $q$. Critically, a single \textbf{cross-group majority vote} is performed over this entire mixed set $\mathcal{O}_{\text{mix}}$ to derive a joint pseudo-label $y_{\text{mix}}^*$. This joint label represents the model's most robust consensus across all variational expressions of the problem.
The advantage for every rollout $o_i \in \mathcal{O}_{\text{mix}}$ is then calculated with respect to this single, consistency-enforcing label $y_{\text{mix}}^*$. The subsequent GRPO update, therefore, optimizes the policy based on a unified, cross-group reward signal, compelling it to become self-consistent across the entire semantic cluster. This mode provides a form of consistency-driven exploration, as it regularizes the policy and ensures its reasoning process is invariant to superficial changes in the problem statement.
\section{Experiments} \label{expeiment}

\begin{table*}[t]

\centering
\small
\setlength{\tabcolsep}{1.7mm}
\begin{tabular}{@{}lllllll@{}}
\toprule
\textbf{Methods} &\textbf{MATH500} &\textbf{AIME2024} &\textbf{AMC2023} &\textbf{GPQA} &\textbf{Average} \\

\midrule
\multicolumn{7}{c}{\textit{RL Post-trained Models \textit{w /} Labeled Data }} \\ \cmidrule{1-6}

DeepSeek-R1-Distill-1.5B (800K) & 52.2 & 2.5 &21.7  & 14.6 & 22.8 \\
\hdashline[1pt/2pt]
\addlinespace[2pt]
DeepSeek-R1-Distill-7B (800K)  & 60.1 & 10.0 & 26.2 & 25.7 & 30.5\ \\
\hdashline[1pt/2pt]
\addlinespace[2pt]
OpenReasoner-Zero-7B (129K) & 79.2 & 13.3 &47.0 &28.4 & 42.0 \\
\hdashline[1pt/2pt]
\addlinespace[2pt]
SimpleRL-Zero-7B  (8.9K)  & 78.2 &26.7 & 60.2 & 27.6 & 48.2 \\

\midrule
\multicolumn{7}{c}{\textit{Qwen3 Family}} \\ \cmidrule{1-6}
Qwen3-1.7B-Instruct  & 59.4 & 3.3 & 28.9 & 13.1 & 26.2 \\
\hdashline[1pt/2pt]
\addlinespace[2pt]
\textit{w /} TTRL  & 73.4\cgreen{(+14.0)} & 15.2\cgreen{(+11.9)} & 54.2\cgreen{(+25.3)} & 22.2\cred{(+9.1)} & 41.3\cgreen{(+15.1)} \\
\rowcolor{lightblue}
\textbf{\textit{w /} TTVS (Ours)}  & \textbf{80.2\cgreen{(+20.8)}} & \textbf{23.3\cred{(+20.0)}} & \textbf{61.4\cgreen{(+32.5)}} & \textbf{26.3  \cgreen{(+16.7)}}
& \textbf{47.5\cgreen{(+21.3)}} \\

\midrule 
Qwen3-4B  & 64.0 & 10.0 & 26.5 & 26.3 & 31.7 \\
\hdashline[1pt/2pt]
\addlinespace[2pt]
\textit{w /} TTRL  & 85.0\cgreen{(+21.0)} & 26.7\cgreen{(+16.7)} & 61.4\cgreen{(+34.9)} &43.0\cred{(+16.7)} & 54.0\cgreen{(+22.3)} \\
\rowcolor{lightblue}
\textbf{\textit{w /} TTVS (Ours)}  &\textbf{90.3\cgreen{(+26.3)}} & \textbf{36.7\cred{(+26.7)}} & \textbf{71.1\cgreen{(+44.6)}} & \textbf{48.9  \cgreen{(+21.7)}} & \textbf{61.5\cgreen{(+29.8)}} \\
\midrule 

Qwen3-8B & 82.2 & 26.9 & 57.8 & 48.0 & 53.7 \\
\hdashline[1pt/2pt]
\addlinespace[2pt]
\textit{w /} TTRL  & 89.2\cgreen{(+7.0)} & 46.7\cgreen{(+19.8)} & 68.6\cgreen{(+10.8)} & 53.0\cred{(+5.0)} & 65.4\cgreen{(+10.7)} \\
\rowcolor{lightblue}
\textbf{\textit{w /} TTVS (Ours)}  & \textbf{92.6\cgreen{(+10.4)}} & \textbf{50.0\cred{(+23.1)}} & \textbf{72.3\cgreen{(+14.5)}} & \textbf{56.1  \cgreen{(+8.1)}} & \textbf{67.8\cgreen{(+14.1)}} \\

\midrule

\multicolumn{7}{c}{\textit{Qwen2.5 Family}} \\ \cmidrule{1-6}
Qwen2.5-Instruct-1.5B & 54.4 & 0.0 & 24.1 & 16.7 & 23.8 \\
\hdashline[1pt/2pt]
\addlinespace[2pt]
\textit{w /} TTRL  & 62.2\cgreen{(+7.8)} & 3.3\cred{(+3.3)} & 32.5\cred{(+8.4)} & 25.3\cgreen{(+8.6)} & 30.8\cgreen{(+7.1)} \\
\rowcolor{lightblue}
\textbf{\textit{w /} TTVS (Ours)}  & \textbf{65.8\cgreen{(+11.4)}} & \textbf{3.3\cred{(+3.3)}} & \textbf{36.1\cgreen{(+12.0)}} & \textbf{28.3  \cgreen{(+11.6)}} & \textbf{33.4\cgreen{(+9.6)}} \\
\midrule 

Qwen2.5-Instruct-3B & 64.2 & 3.3 & 32.5 & 29.3 & 32.3 \\
\hdashline[1pt/2pt]
\addlinespace[2pt]
\textit{w /} TTRL   & 71.2\cgreen{(+7.0)} & 6.7\cgreen{(+3.4)} & 39.8\cgreen{(+7.3)} & 33.3\cred{(+4.0)} & 37.8\cgreen{(+5.4)} \\
\rowcolor{lightblue}
\textbf{\textit{w /} TTVS (Ours)}  & \textbf{75.0\cgreen{(+10.8)}} & \textbf{6.7\cred{(+3.4)}} & \textbf{42.2\cgreen{(+9.7)}} & \textbf{36.4  \cgreen{(+7.1)}} & \textbf{40.1\cgreen{(+7.8)}} \\
\midrule 

Qwen2.5-Instruct-7B & 70.4 & 10.2 & 41.0 & 36.4 & 39.5 \\
\hdashline[1pt/2pt]
\addlinespace[2pt]
\textit{w /} TTRL   & 77.6\cgreen{(+7.2)} & 13.3\cgreen{(+3.1)} & 44.6\cgreen{(+3.6)} & 48.5\cred{(+12.1)} & 46.0\cgreen{(+6.5)} \\
\rowcolor{lightblue}
\textbf{\textit{w /} TTVS (Ours)}  & \textbf{79.4\cgreen{(+9.0)}} & \textbf{16.7\cred{(+6.5)}} & \textbf{47.2\cgreen{(+6.2)}} & \textbf{51.5  \cgreen{(+15.1)}} & \textbf{48.7\cgreen{(+9.2)}} \\

\midrule

\multicolumn{7}{c}{\textit{LLaMA Family}} \\ \cmidrule{1-7}
LLaMA-3.2-3B-Instruct & 40.8 & 0.8 & 15.1 & 12.5 & 17.3 \\
\hdashline[1pt/2pt]
\addlinespace[2pt]
\textit{w /} TTRL   & 51.0\cgreen{(+10.2)} & 3.3\cgreen{(+2.5)} & 25.3\cgreen{(+10.2)} & 16.1\cred{(+3.6)} & 23.9\cgreen{(+6.6)} \\
\rowcolor{lightblue}
\textbf{\textit{w /} TTVS (Ours)}  & \textbf{54.1\cgreen{(+13.3)}} & \textbf{6.9\cred{(+6.1)}} & \textbf{27.9\cgreen{(+12.8)}} & \textbf{17.7  \cgreen{(+5.2)}} & \textbf{26.7\cgreen{(+9.4)}} \\

\midrule 
LLaMA3.1-8B & 48.6 & 4.6 & 23.3 & 30.8 & 26.8 \\
\hdashline[1pt/2pt]
\addlinespace[2pt]
\textit{w /} TTRL   & 63.7\cgreen{(+15.1)} & 10.0\cgreen{(+5.4)} & 32.3\cgreen{(+9.0)} & 34.1\cred{(+3.3)} & 35.0\cgreen{(+8.2)} \\
\rowcolor{lightblue}
\textbf{\textit{w /} TTVS (Ours)}  & \textbf{64.9\cgreen{(+16.3)}} & \textbf{10.0\cred{(+5.4)}} & \textbf{33.5\cgreen{(+10.2)}} & \textbf{37.3 \cgreen{(+6.5)}} & \textbf{36.4\cgreen{(+9.6)}} \\

\bottomrule
\end{tabular}
\caption{Performance comparison of TTVS with state-of-the-art methods on various model architectures. For RL Post-trained Models \textit{w /} Labeled Data ($\cdot$), ($\cdot$) indicates the amount of labeled data used during post-training. 
\label{tab:main}
}
\vspace{-5mm}
\end{table*}

\subsection{Experimental Setup}
\textbf{Datasets.} 
We evaluate model performance on a wide range of mathematical reasoning benchmarks. These include MATH-500 \cite{hendrycks2021math500}, a curated collection of 500 competition-level problems from the MATH dataset;  AIME-2024 \cite{li2024aime_amc}, comprising challenging problems from the 2024 American Invitational Mathematics Examination; AMC-2023 \cite{li2024aime_amc}, a dataset sourced from the American Mathematics Competitions that contains 83 samples comprising a series of progressively challenging mathematical tests designed for middle and high school students; and GPQA \cite{rein2024gpqa}, a high-quality and exceptionally challenging subset of the Graduate-Level Google-Proof Question Answering benchmark that curated from domains such as physics, chemistry, and biology.

\textbf{Models.} To assess the generalizability of our proposed method, we conducted extensive experiments across three distinct model families, encompassing a wide range of parameter scales.  Specifically, the models we experiment with are as follows. \textit{1) Qwen3 Family} \cite{yang2025qwen3}: Qwen3-1.7B-Instruct, Qwen3-4B, and Qwen3-8B. \textit{2)Qwen2.5 Family} \cite{yang2024qwen2}: Qwen2.5-Instruct-1.5B, Qwen2.5-Instruct-3B, and Qwen2.5-Instruct-7B. \textit{3) LLaMA Family} \cite{dubey2024llama}: LLaMA-3.2-3B-Instruct and LLaMA3.1-8B.

\textbf{Baselines.} We compare our method with the following methods: 1) RL Post-Trained Models: they are leading models with architectures similar to our own, which have undergone extensive reinforcement learning on large-scale labeled data, including DeepSeek-R1-Distill (1.5B \& 7B) \cite{guo2025deepseek}, SimpleRL-Zero-7B \cite{zeng2025simplerl}, and OpenReasoner-Zero-7B \cite{hu2025openreasoner}. It is worth noting that our TTVS operates under a test-time adaptation paradigm using only unlabeled data, which represents a fundamental difference from these methods.  2) TTRL \cite{zuo2025ttrl}: similar to our TTVS, this method also utilizes test-time reinforcement learning; however, it only estimates rewards through majority voting on original unlabeled data during training.

\textbf{Implementation Details.} 
For the implementation of TTVS, we employ GRPO as the optimization strategy across all benchmarks. The policy model is optimized using the AdamW optimizer with a cosine learning rate schedule, peaking at $5 \times 10^{-7}$. During the rollout phase, 32 responses are sampled (temperature = 0.6) for the voting-based label estimation. By default, we set $L_{max}$ is 1024, $\tau_{\text{low}}$ is 0.125 and $\tau_{\text{high}}$ equals 0.875. To ensure a fair comparison with the TTRL, the maximum generation length is constrained to 3072 tokens. 
For evaluation metrics, we evaluate model performance using pass@1 Score. For each problem, we generate 16 candidate responses using temperature sampling (temperature = 0.6, top-p = 0.95) and assess the correctness of the top-ranked solution.
Moreover, we set the warmup steps as 40 for IGE and 60 for CGE on aross four reasoning benchmarks. The number of training episodes was configured to 10, 10, 20, and 40 for MATH500, GPQA, AMC2023 and AIME2024, scaled according to their size. All experiments were conducted using 8 NVIDIA GeForce H800 80GB GPUs.

\subsection{Main Results}

To illustrate the efficacy of our proposed method, we report the performance comparison of TTVS against several state-of-the-art (SOTA) methods. Our evaluation spans 8 models across 3 distinct model families, with detailed results presented in Table~\ref{tab:main}. 
\textbf{1) Comparison with RL Post-trained Models with Labeled Data}, TTVS outperforms these methods on most settings. When applied to Qwen2.5-Instruct-7B, TTVS exhibits a clear performance advantage over leading 7B models DeepSeek-R1-Distill-7B, OpenReasoner-Zero-7B, particularly on MATH500 and GPQA benchmark. Furthermore, the efficiency of TTVS is underscored by the results that even our smaller Qwen2.5-Instruct-1.5B model outperforms the larger, post-trained DeepSeek-R1-Distill-7B (e.g., 33.4\% vs. 30.5\% average score). This indicates that TTVS can unlock latent reasoning abilities more effectively using test-time self-improvement. 
\textbf{2) Compared to the SOTA test-time RL method (TTRL)}, our TTVS demonstrates improved performance across various benchmarks and model types. Notably, on Qwen3-4B, TTVS yields absolute gains of 5.3\% (MATH500), 10.0\% (AIME2024),  9.7\% (AMC2023), and 5.9\% (GPQA).  These results demonstrate the effectiveness of TTVS and highlight the potential of self-improvements to facilitate greater exploration using online variational synthesis.

\begin{table}[t]
\centering
\resizebox{0.435\textwidth}{!}{
\begin{tabular}{@{}lccc@{}}
\toprule
\textbf{Methods}
& \textbf{MATH500} 
& \textbf{AMC2023} 
& \textbf{GPQA}  \\ 
\midrule
Qwen3-4B &64.0 &26.5 &26.3  \\
\hdashline[1pt/2pt]
$\hookrightarrow$\textbf{CGE} &86.4 &62.6 &43.9  \\
$\hookrightarrow$\textbf{IGE} &88.4 &69.8 &46.6  \\
\hdashline[1pt/2pt]
$\hookrightarrow$TTVS &90.3 &71.1 &48.0 \\
\bottomrule
\end{tabular}}
\caption{Performance analysis of different components in TTVS.}
\vspace{-2mm}
\label{tab:components}
\end{table}

\begin{figure}[!t]
    \center
    \includegraphics[width=0.40\textwidth]{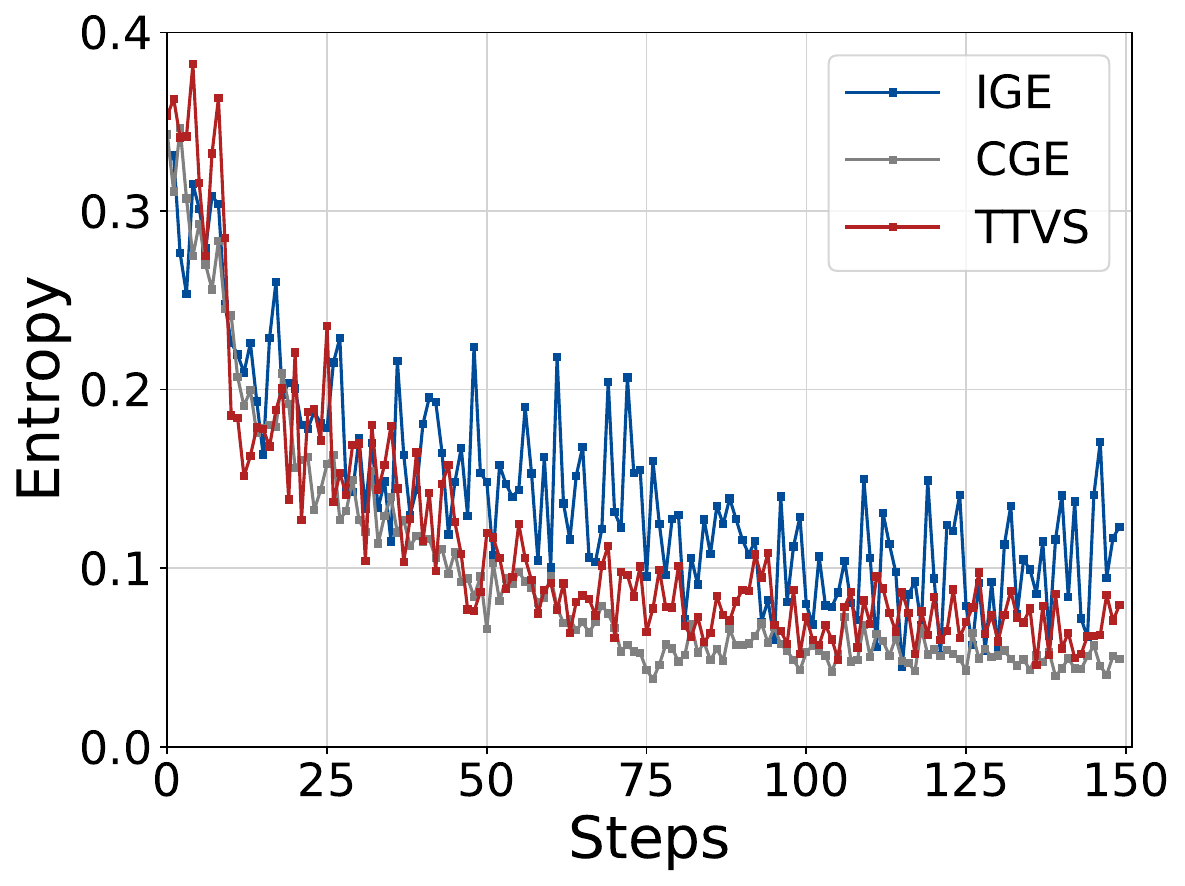}
    \caption{
    Entropy Curve of TTVS components on AMC2023 using Qwen3-4B.
    }
    \label{fig:compoenet}
    \vspace{-3mm}
\end{figure}

\begin{table}[t]
\centering
\begin{spacing}{1.2}
\resizebox{0.48\textwidth}{!}{
\begin{tabular}{@{}lccc@{}}
\toprule
\textbf{Methods}
& \textbf{MATH500} 
& \textbf{AIME2024} 
& \textbf{AMC2023}  \\ 
\midrule
Qwen3-4B &64.0 &10.0  &26.5  \\
\hdashline[1pt/2pt]
$\hookrightarrow$TTVS (\textit{w/} OPO) &89.6 &33.3 &69.8  \\
\hdashline[1pt/2pt]
$\hookrightarrow$TTVS (\textit{w/} DAPO) &88.8 &36.7 &71.6  \\
\hdashline[1pt/2pt]
$\hookrightarrow$TTVS (\textit{w/} GRPO) &90.3 &36.7 &71.1 \\
\bottomrule
\end{tabular}}
\end{spacing}
\caption{Performance comparison of different RL algorithms for TTVS using Qwen3-4B.}
\vspace{-3mm}
\label{tab:rlcompatibility}
\end{table}

\subsection{Ablation Studies}
\subsubsection{Effectiveness of Components}
To validate the contributions of the individual components within our TTVS, we conduct ablation studies on Qwen3-4B. As shown in Table \ref{tab:components}, we systematically evaluate the performance of several variants across the MATH500, AMC2023, and GPQA benchmarks. The naive Qwen3-4B is established as the baseline, and we directly perform the inference on various reasoning tasks and have poor performance.  We then evaluate two key components of our framework in isolation.  First, the Consistency-Guided Exploitation (CGE) module yields substantial performance gains over the baseline (e.g., 86.4\% v.s. 64.0\% on MATH500). Second, the Intra-group Exploitation (IGE) module achieves even greater improvements, with scores of 88.4\%, 69.8\%, and 46.6\%. It obtains further performance improvements and exhibits persistently high entropy throughout training in Figure \ref{fig:compoenet}. Finally, the full TTVS framework integrates both CGE and IGE, leveraging their complementary strengths to balance generalization and accuracy. This synergistic combination achieves more steady exploration and the optimal performance across all benchmarks, reaching 90.3\% on MATH500, 71.1\% on AMC2023, and 48.0\% on GPQA, thereby confirming the efficacy of our synergistic exploration.
\begin{figure*}[!t]
    \center
    \includegraphics[width=0.79\textwidth]{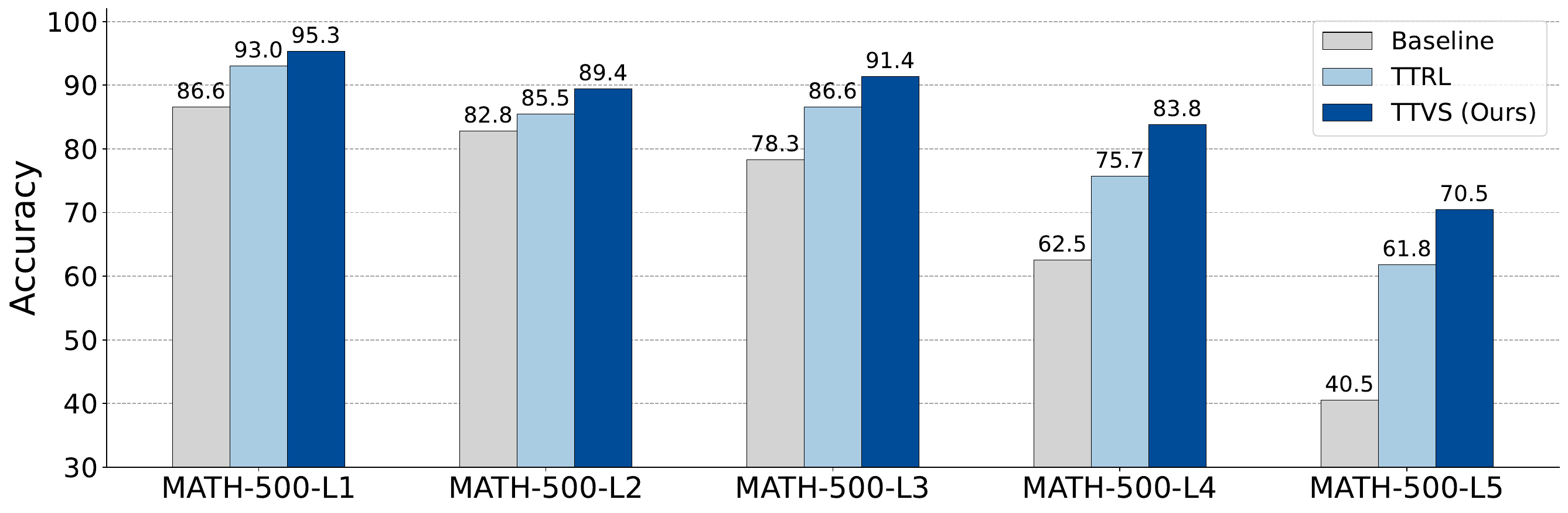}
    \caption{
    Comparative performance analysis across the five difficulty levels of MATH-500.
    }
    \label{fig:math1_5}
\end{figure*}
\subsubsection{Impacts of Different RL Algorithms}
To assess the versatility of our TTVS, we evaluated its performance when integrated with three distinct reinforcement learning algorithms: GRPO \cite{shao2024grpo}, OPO \cite{hao2025opo}, and DAPO \cite{yu2025dapo}. 
GRPO normalizes rewards within a group of sampled responses to calculate a relative advantage. OPO employs a length-weighted average as an optimal reward baseline to stabilize training. DAPO incorporates a suite of techniques, such as clip-higher and a token-mean objective, to enhance stability in large-scale training. 
As shown in Table \ref{tab:rlcompatibility}, TTVS consistently delivers substantial performance improvements over the Qwen3-4B (baseline), regardless of the specific RL optimizer used. Furthermore, the performance of the three RL algorithms was closely aligned, with only minor differences observed. GRPO achieved the highest score on MATH500 (90.3\%), while DAPO and GRPO performed identically on AIME2024 (36.7\%). This consistent high performance validates the robustness of TTVS and confirms its compatibility with a range of policy optimization strategies.

\subsubsection{Computational Cost Analysis}
\begin{figure}[!t]
    \center
    \includegraphics[width=0.50\textwidth]{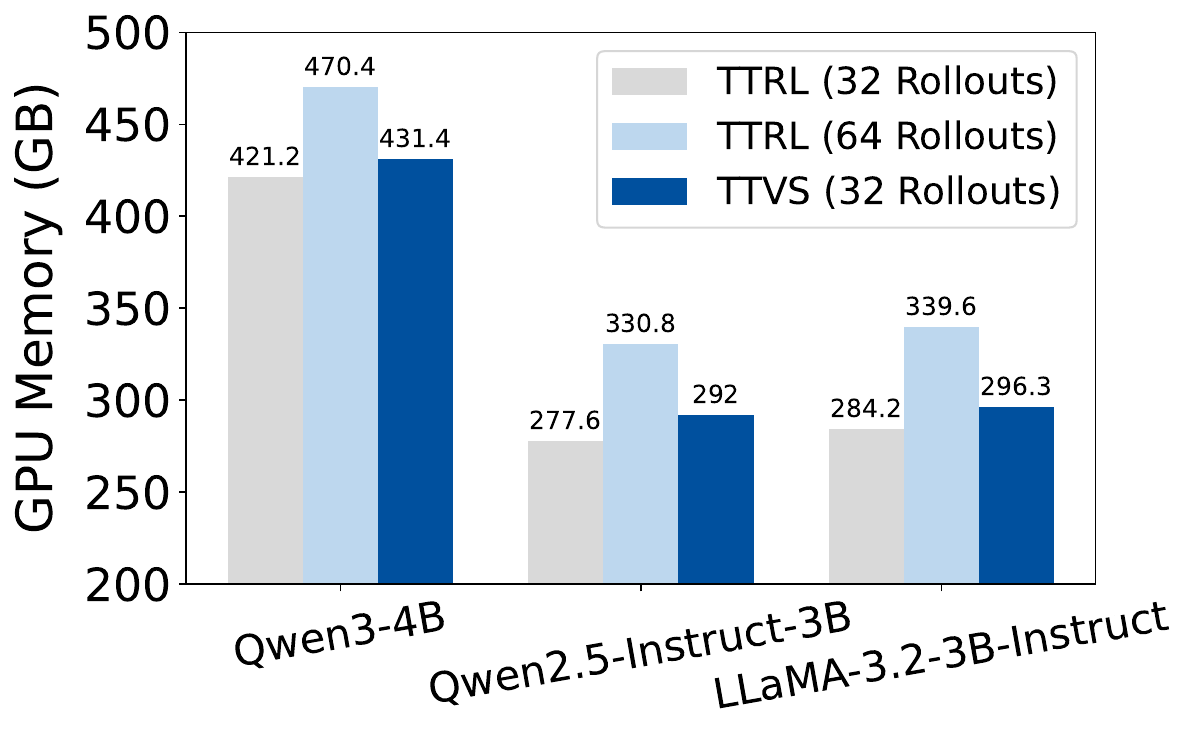}
    \caption{
    Computational cost during test-time training on Qwen3-4B
    }
    \label{fig:training_cost}
    \vspace{-2mm}
\end{figure}

Finally, we conducted a detailed quantitative analysis of the resources used during both test-time training and the inference phase. \textbf{1) During Test-Time Training}, as shown in Figure \ref{fig:training_cost}, the GPU memory consumption of TTVS is nearly identical to the TTRL baseline phase with the same number of rollouts (32). Even when compared to a TTRL baseline with double the rollouts (64), our TTVS requires substantially less GPU memory while still achieving superior accuracy. This confirms that our method achieves its significant performance improvements without incurring substantial additional computational overhead during test-time training. \textbf{2) At the inference phase}, the model produced by TTVS is also more efficient. As shown in Table \ref{tab:inference_cost}, TTVS generates more concise answers (shorter token length), leading to a slightly lower memory footprint. Overall, these results demonstrate that TTVS offers a superior trade-off between performance and computational cost. Its advantages are derived from the quality of its dynamic data synthesis, not from a greater quantity of computation. We provide more clarification on TTVS's computational trade-off in Appendix \ref{text: clarification}.

\begin{table}[t]
\centering
\begin{spacing}{1.3}
\resizebox{0.48\textwidth}{!}{
\begin{tabular}{@{}lccc@{}}
\toprule
\textbf{Methods}
& \textbf{GPU Memory (GB)} 
& \textbf{Token Length} 
& \textbf{Throughput (Token/s)}  \\ 
\midrule
TTRL &401.4	&1925.5	&15.60  \\
\hdashline[1pt/2pt]
\textbf{TTVS} &398.0	&1864.9	&14.28 \\
\bottomrule
\end{tabular}}
\end{spacing}
\caption{Computational cost at inference phase based on Qwen3-4B using 32 rollouts}
\vspace{-3mm}
\label{tab:inference_cost}
\end{table}

\subsubsection{Impacts of Question Difficulty Levels}
To validate the impacts of question difficulty levels for TTVS, we conducted a fine-grained analysis using the MATH-500 dataset. The dataset was partitioned into five subsets corresponding to its annotated difficulty levels (L1 to L5). We then evaluated the performance of the baseline (Qwen3-4B), TTRL, and our TTVS framework on each subset. As illustrated in Figure \ref{fig:math1_5},  it shows crucial trend: the performance gap between TTVS and TTRL widens as the complexity of the problems increases. On Level 1 problems, TTVS holds a slight edge over TTRL (95.3\% vs. 93.0\%). This advantage becomes far more substantial on Level 5 problems, where TTVS achieves an accuracy of 69.5\%—with 8.7\% and 30.0\% performance improvements over TTRL  the baseline, respectively. These results demonstrate the superior robustness of TTVS, suggesting that its variational synthesis mechanism effectively compensates for these knowledge gaps, enabling the model to learn and reason more effectively on challenging tasks. We provide more experimental results in Appendix.
\section{Conclusion} \label{conclusion}
In this paper, we introduced Test-Time Variational Synthesis (TTVS), a novel framework designed to enhance the self-exploration capabilities of Large Reasoning Models (LRMs) through dynamic data augmentation. TTVS facilitates the online generation of semantically consistent synthetic variations of test queries, thereby enriching the data without the need for additional human annotations. This augmented dataset then enables a hybrid exploration strategy for optimizing the policy model during test-time. Comprehensive experiments demonstrate that TTVS achieves superior performance compared to existing test-time adaptation methods and state-of-the-art RL post-training strategies across a diverse range of mathematical reasoning benchmarks and model architectures, establishing a new performance benchmark for test-time reinforcement learning.
\section*{Limitations}\label{07limitation}
While the proposed TTVS method demonstrates strong performance across a range of reasoning benchmarks and model architectures, the present study is subject to several limitations that open avenues for future investigation.
First, due to computational constraints, our primary experiments cannot be conducted on some larger-scale LLMs, such as Qwen2.5-32B-Instruct, Qwen3-32B. Validating the scalability and effectiveness of TTVS on such models is an important direction for future work. Second, the scope of this study is confined to language-only models. Consequently, the efficacy of our approach in multimodal architectures remains unexplored. Investigating whether our method can be adapted to achieve competitive performance on test-time vision-language tasks constitutes a promising area for further research.


\bibliography{acl}
\newpage

\appendix

\quad
\newpage

\begin{table}[t]
\centering

\begin{spacing}{1.3}
\resizebox{0.5\textwidth}{!}{
\begin{tabular}{@{}lccc@{}}
\toprule
\multicolumn{2}{c}{\textbf{\bskblue{Filtering thresholds}}}
& \textbf{\bskblue{Qwen3-4B}} 
& \textbf{\bskblue{Qwen2.5-Instruct-3B}}  \\ 
\midrule
\bskblue{$\tau_{low}=0.0$} &\bskblue{$\tau_{high}=0.875$}	&\bskblue{86.4}	&\bskblue{72.0} \\
\bskblue{$\tau_{low}=0.125$} &\bskblue{$\tau_{high}=0.875$}	&\bskblue{\textbf{90.3}} &\bskblue{\textbf{75.0}} \\
\bskblue{$\tau_{low}=0.25$} &\bskblue{$\tau_{high}=0.875$}	&\bskblue{86.2}	&\bskblue{72.4} \\
\bskblue{$\tau_{low}=0.125$} &\bskblue{$\tau_{high}=0.625$}	&\bskblue{88.4}	&\bskblue{73.8} \\
\bskblue{$\tau_{low}=0.125$} &\bskblue{$\tau_{high}=1.0$}	&\bskblue{88.7}	&\bskblue{73.2} \\
\bottomrule
\end{tabular}}
\end{spacing}
\caption{\bskblue{Ablation study for filtering thresholds}}
\vspace{-1mm}
\label{tab:filter_thres}
\end{table}

\begin{table}[t]
\centering
\resizebox{0.45\textwidth}{!}{
\begin{tabular}{@{}lcc@{}}
\toprule
\textbf{\bskblue{Methods}}
& \textbf{\bskblue{Qwen3-4B}} 
& \textbf{\bskblue{Qwen2.5-Instruct-3B}} \\ 
\midrule
\bskblue{512} &\bskblue{86.1}	&\bskblue{72.2}  \\
\bskblue{1024} &\bskblue{90.3}	&\bskblue{75.0} \\
\bskblue{2048}	&\bskblue{90.7}	&\bskblue{74.9} \\
\bottomrule
\end{tabular}}
\caption{\bskblue{Ablation study for Maximum Token Length.}}
\vspace{-3mm}
\label{tab:L_max}
\end{table}

\section{Appendix} \label{appendix}
\bskblue{\subsection{More Details of the Online Filtering Mechanism} \label{online_filtering}
This mechanism is a crucial two-step process designed to curate a training batch of suitable difficulty and quality without verifiable labels. 
\textbf{Step 1: Difficulty Scaffolding (on Original Query $q$)}: To assess difficulty, we first generate a pseudo-label $y*$ for the original query $q$ via majority voting. We then calculate the initial group accuracy, $acc(q)$, as the fraction of candidate answers matching this pseudo-label (Equation 10). This $acc(q)$ metric serves as a proxy for the problem's difficulty for the current model. We only proceed with synthesis if this accuracy falls within a predefined range $[\tau_{low}, \tau_{high}]$, ensuring the model focuses on problems that are neither trivial nor intractable.
\textbf{Step 2: Quality Control (on Synthesized Query $q'$)}: To prevent the introduction of malformed or unsolvable problems, we apply a simple but effective heuristic: each newly synthesized query $q'$ must not exceed a maximum token length $L_{max}$. This filters out incomplete or overly complex generations. The effectiveness of our entire filtering mechanism is verified by a series of ablation studies in terms of filtering thresholds $\tau_{low}$, $\tau_{high}$, and $L_{max}$.}

\begin{figure}[t]
  \centering
  
   \subfloat[MATH500 and AMC2023. ]{\includegraphics[width=0.23\textwidth]{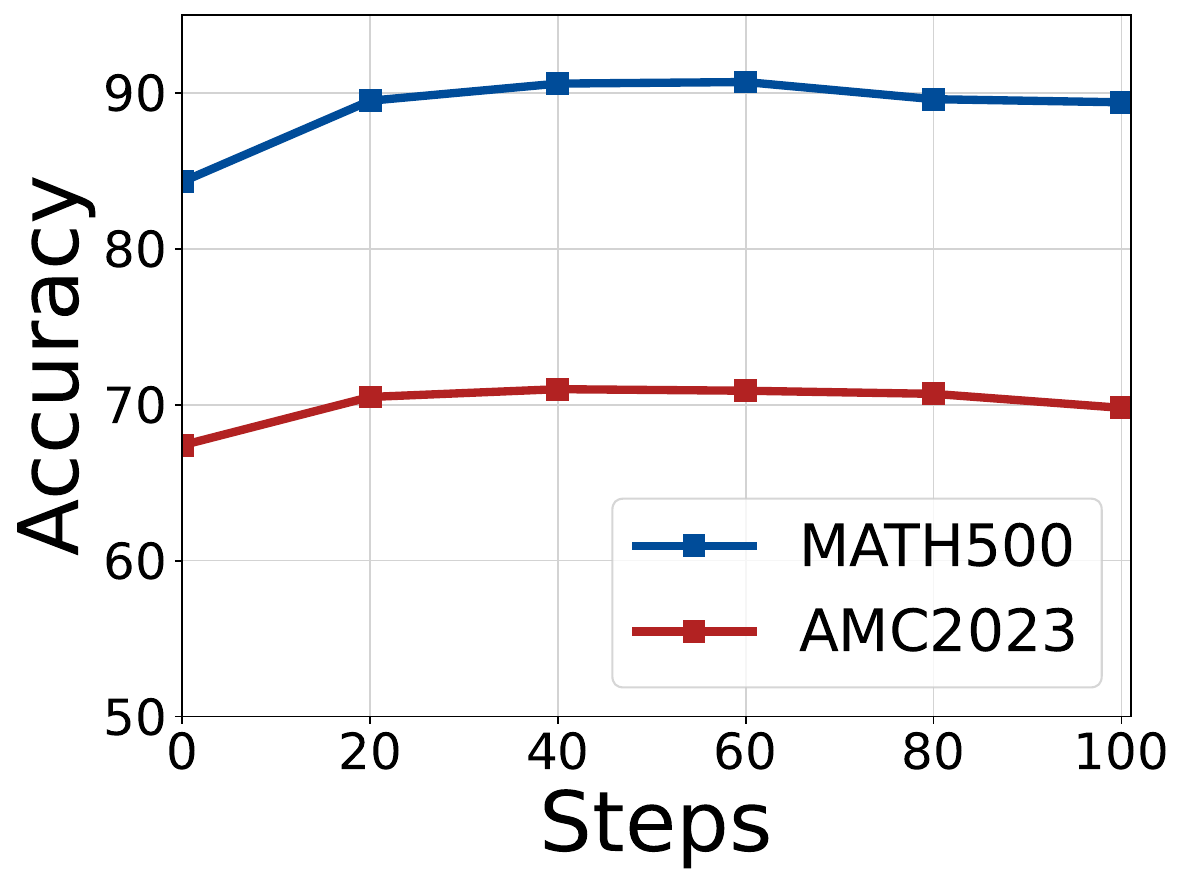}
   \label{subfig:intra_a}}\hfill
   \subfloat[GPQA and AIME2024.]{\includegraphics[width=0.23\textwidth]{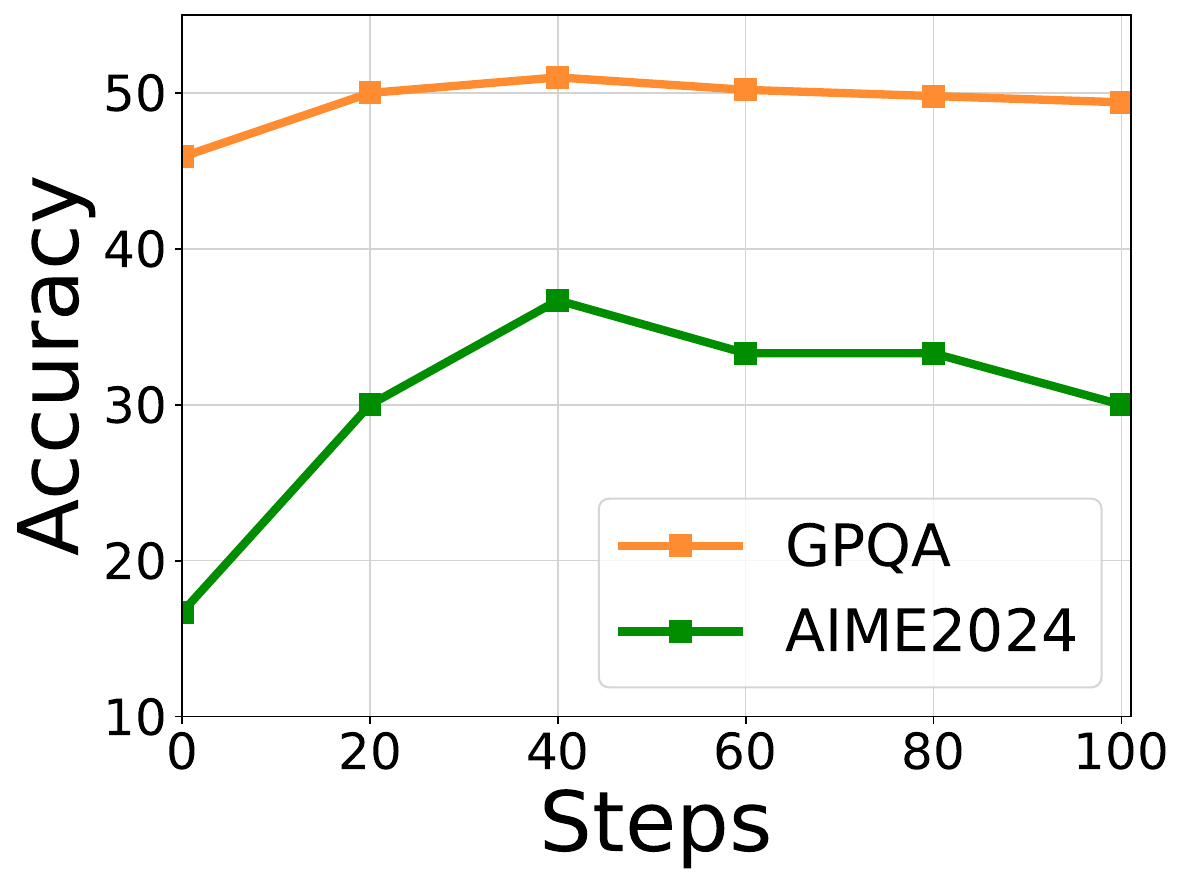}
    \label{subfig:intra_b}}\hfill
  \caption{ Hyperparameter analysis of the warmup steps for intra-group exploration ($E_{intra}$) on various reasoning datasets..  }
  \label{fig:hyperparameters}
\end{figure}

\begin{figure}[t]
  \centering
    \subfloat[MATH500 and AMC2023.]{\includegraphics[width=0.23\textwidth]{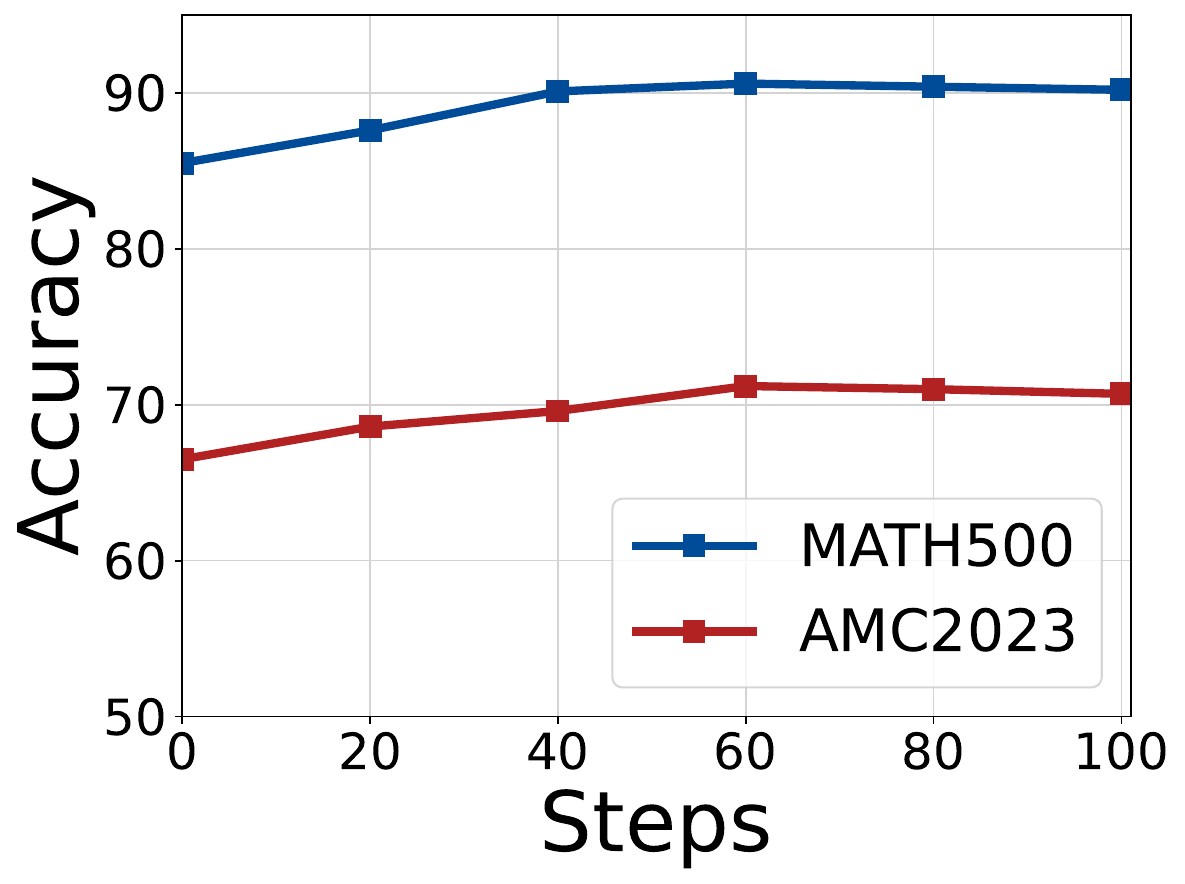}
   \label{subfig:crpss_a}}\hfill
   \subfloat[GPQA and AIME2024.]{\includegraphics[width=0.23\textwidth]{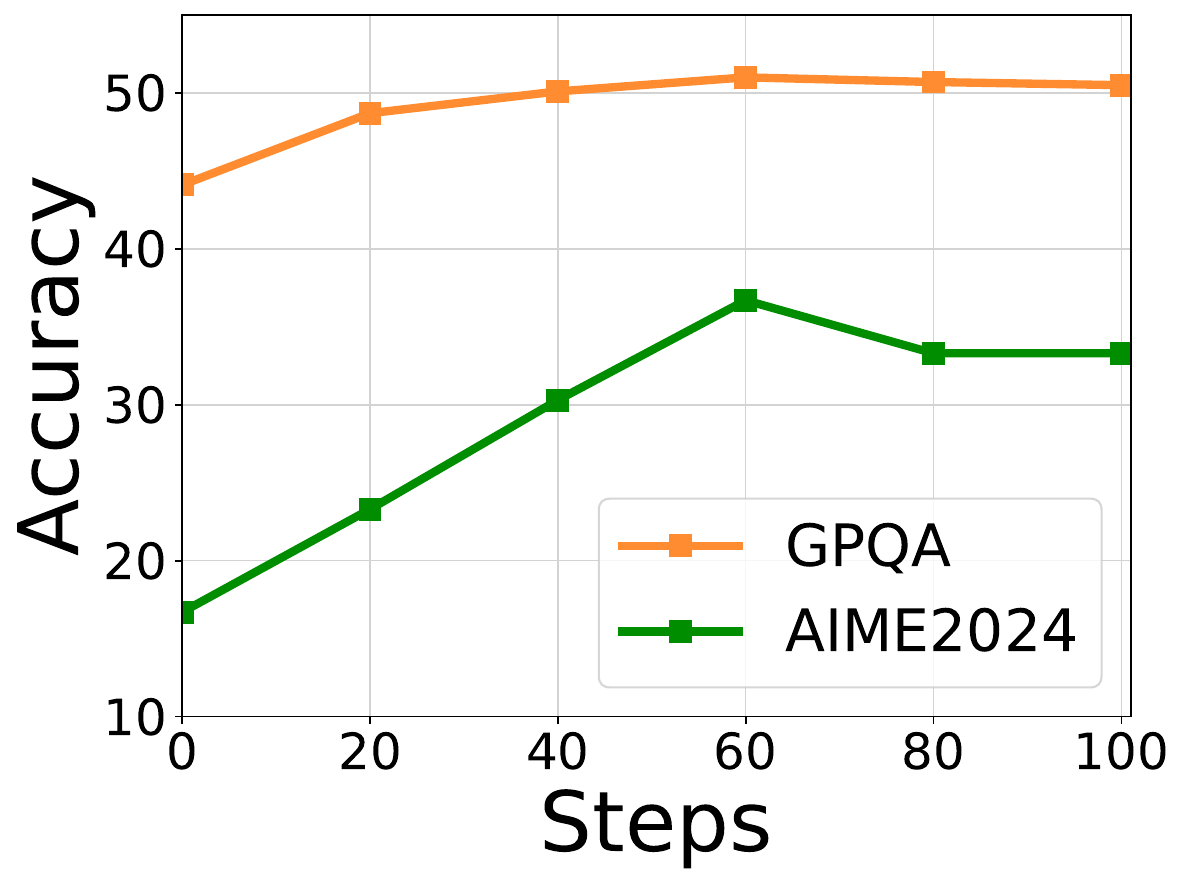}
    \label{subfig:cross_b}}\hfill
  \caption{\bskorange{ Hyperparameter analysis of the warmup steps cross-group exploration ($E_{cross}$) on various reasoning datasets.}}
  \label{fig:hyperparameters}
\end{figure}

\bskblue{\subsection{Ablation Study for Hyperparameters}\label{text: hyperparameters}
We first investigate the sensitivity to filtering thresholds $\tau_{low}$ and $\tau_{high}$ on MATH500 dataset. Table \ref{tab:filter_thres} shows that our chosen thresholds ($\tau_{low}=0.125$, $\tau_{high}=0.875$) achieve optimal performance. We have also conducted an additional ablation study on $L_{max}$ to further validate our quality control step. The results in Table \ref{tab:L_max} show that $L_{max}=1024$ provides the best balance, effectively filtering out malformed queries without being overly restrictive. The stable performance around this value further demonstrates the robustness of our filtering approach.}
\bskorange{Furthermore, our TTVS incorporates other critical hyperparameters: the number of warm-up steps for intra-group exploration ($E_{intra}$) and for cross-group exploration ($E_{cross}$). The former determines the transition point from exploration to accuracy-driven exploitation, while the latter controls the onset of consistency-driven exploration. To identify the optimal values for these parameters, we conducted a series of sensitivity analyses. First, we evaluated the impact of $E_{intra}$ on model performance across a range of values: {0, 20, 40, 60, 80, 100}. As illustrated in Figures \ref{subfig:intra_a} and \ref{subfig:intra_b}, performance generally peaked when $E_{intra}$ was set to 40 across most reasoning tasks, including AMC2023, GPQA, and AIME2024. Subsequently, we investigated the effect of $E_{cross}$ in a similar manner. The results, shown in Figures \ref{subfig:crpss_a} and \ref{subfig:cross_b}, indicate that the TTVS framework achieved optimal or near-optimal performance on all benchmarks when $E_{cross}$ was set to 60. Consequently, we adopted default values of 40 and 60 for $E_{intra}$ and $E_{cross}$ in all experiments unless otherwise specified. We also analyzed the sensitivity to our difficulty filtering thresholds. }
\bskblue{\subsection{Ablation Study for Rollouts} \label{rollouts}
We first conducted an ablation study where we doubled the computational budget for TTRL by increasing its rollouts to 64. As shown in Table \ref{tab:budget}, simply increasing computation for TTRL yields only marginal performance gains (2\%-4\%), demonstrating diminishing returns on a static dataset. More importantly, our more compute-efficient TTVS (with 32 rollouts) still significantly outperforms the more expensive TTRL (with 64 rollouts). Furthermore, we investigated the impact of varying the number of rollouts for TTVS. As shown in Table \ref{tab:rollouts}, TTVS effectively improves performance with more variants. This indicates that our chosen value offers a strong balance between performance and efficiency, and the method is not overly sensitive to this parameter.}

\begin{table}[t]
\centering
\resizebox{0.435\textwidth}{!}{
\begin{tabular}{@{}lccc@{}}
\toprule
\textbf{\bskblue{Methods}}
& \textbf{\bskblue{MATH500}} 
& \textbf{\bskblue{AMC2023}} 
& \textbf{\bskblue{GPQA}}  \\ 
\midrule
\bskblue{Qwen3-4B} &\bskblue{64.0} &\bskblue{26.5} &\bskblue{26.3}  \\
\hdashline[1pt/2pt]
\bskblue{TTRL (32 Rollouts)} &\bskblue{85.0} &\bskblue{61.4}	&\bskblue{43.0}  \\
\bskblue{TTRL (64 Rollouts)} &\bskblue{86.9} &\bskblue{65.5}	&\bskblue{45.0}  \\
\bskblue{TTVS (32 Rollouts)} &\bskblue{90.3} &\bskblue{71.1} &\bskblue{48.0} \\
\bottomrule
\end{tabular}}
\caption{\bskblue{Performance comparison with increased rollouts for TTRL}}
\vspace{-2mm}
\label{tab:budget}
\end{table}

\begin{table}[t]
\centering
\resizebox{0.435\textwidth}{!}{
\begin{tabular}{@{}lccc@{}}
\toprule
\textbf{\bskblue{Methods}}
& \textbf{\bskblue{Qwen3-4B}}
& \textbf{\bskblue{Qwen2.5-Instruct-3B}}   \\ 
\midrule
\bskblue{TTVS (16 Rollouts)} &\bskblue{88.0} &\bskblue{73.8}  \\
\bskblue{TTVS (32 Rollouts)} &\bskblue{90.3} &\bskblue{75.0}  \\
\bskblue{TTVS (64 Rollouts)} &\bskblue{92.9} &\bskblue{76.2} \\
\bottomrule
\end{tabular}}
\caption{\bskblue{Ablation study for the number of rollouts for TTVS.}}
\vspace{-2mm}
\label{tab:rollouts}
\end{table}

\bskblue{\subsection{Clarification for Trade-off Structure} \label{text: clarification}
To analyze computational cost, we provide a detailed methodology clarification demonstrating that TTVS's performance gains stem from its novel mechanism, not from a higher computational budget. TTVS was designed for efficiency and does not use $k$ times more rollouts in parallel. For each optimization step of the policy model, the number of rollouts used is identical to the baseline TTRL (i.e., 32 rollouts). TTVS sequentially utilizes rollouts from different data sources (original, synthetic, mixed) to improve the quality and diversity of the training signal, rather than increasing the quantity of parallel computation.}

\bskblue{\subsection{Evaluation on More Unseen Dataset}\label{text: unseen_data}
To address the potential for data contamination, we conducted performance analysis in Table \ref{tab:deepmath103k} on a subset of the recently released DeepMath-103K dataset \cite{he2025deepmath103k}, which is unseen by our models. We created a representative subset by randomly sampling 20 question-answer pairs from each difficulty level. TTVS consistently outperforms both the initial model and the TTRL baseline, demonstrating that its gains are genuine and not an artifact of data leakage.}
\begin{table}[t]
\centering
\resizebox{0.435\textwidth}{!}{
\begin{tabular}{@{}lccc@{}}
\toprule
\textbf{\bskblue{Methods}}
& \textbf{\bskblue{LLaMA-3.2-3B-Instruct}} 
& \textbf{\bskblue{Qwen3-4B}}   \\ 
\midrule
\bskblue{Init} &\bskblue{44.1} &\bskblue{53.2}  \\
\bskblue{TTRL} &\bskblue{46.6} &\bskblue{62.8}  \\
\bskblue{TTVS} &\bskblue{49.0} &\bskblue{65.9}  \\
\bottomrule
\end{tabular}}
\caption{\bskblue{Results on DeepMath-103K subset.}}
\vspace{-2mm}
\label{tab:deepmath103k}
\end{table}
\begin{table}[t]
\centering
\resizebox{0.435\textwidth}{!}{
\begin{tabular}{@{}lccc@{}}
\toprule
\textbf{\bskblue{Methods}}
& \textbf{\bskblue{LLaMA-3.2-3B-Instruct}} 
& \textbf{\bskblue{Qwen3-4B}}   \\ 
\midrule
\bskblue{Init} &\bskblue{49.8} &\bskblue{56.6}  \\
\bskblue{TTRL} &\bskblue{51.3} &\bskblue{58.1}  \\
\bskblue{TTVS} &\bskblue{52.1} &\bskblue{59.1} \\
\bottomrule
\end{tabular}}
\caption{\bskblue{Results on MedMCQA.}}
\vspace{-2mm}
\label{tab:medmcqa}
\end{table}
\bskblue{\subsection{Evaluation on Domain-Specific Dataset}\label{text: domain_data}
To investigate the practical value of TTVS in domains with scarce annotated data, we evaluated it on the well-known medical benchmark, MedMCQA \cite{pal2022medmcqa}. As shown in Table \ref{tab:medmcqa} TTVS again shows a clear and consistent advantage over TTRL, confirming its effectiveness in specialized domains.}

\bskblue{\subsection{Performance on Base Models without Instruction-following Ability} \label{text:base_model}
To demonstrate that the benefits of TTVS are not limited to instruction-following models, we conducted an experiment on the Qwen2.5-Math-1.5B (Base model). To ensure this condition is met even for base models with initial limitations, we introduced an effective expert-level model warmup phase. For the first 100 training steps, we use a small set of high-quality synthetic data generated by an expert model (DeepSeek R1). It is designed to efficiently bootstrap the base model's specific rewriting skills, enabling it to fully participate in the main TTVS self-improvement loop. Our results in Table \ref{tab:basemodel} demonstrate that with this minimal warmup, TTVS significantly outperforms TTRL on the same base model. This confirms that the core benefits of TTVS—harnessing dynamic data synthesis for self-improvement—are not restricted to models with strong pre-existing capabilities. Our warmup strategy provides a simple yet powerful mechanism to unlock the potential of TTVS on a broader range of models, showcasing the versatility of our framework.}

\begin{table}[t]
\centering

\resizebox{0.435\textwidth}{!}{
\begin{tabular}{@{}lccc@{}}
\toprule
\textbf{\bskblue{Methods}}
& \textbf{\bskblue{MATH500}} 
& \textbf{\bskblue{AMC2023}}
& \textbf{\bskblue{GPQA}}  \\ 
\midrule
\bskblue{Qwen2.5-Math-1.5 }&\bskblue{32.7}	&\bskblue{28.6} &\bskblue{24.9}  \\
\hdashline[1pt/2pt]
\bskblue{TTRL} &\bskblue{70.7} &\bskblue{48.9}	&\bskblue{26.7}  \\
\bskblue{TTVS} &\bskblue{76.0} &\bskblue{50.4}	&\bskblue{28.6} \\
\bottomrule
\end{tabular}}
\caption{\bskblue{Performance Analysis on Base Model.}}
\vspace{-2mm}
\label{tab:basemodel}
\end{table}

\begin{table}[t]
\centering
\resizebox{0.435\textwidth}{!}{
\begin{tabular}{@{}lccc@{}}
\toprule
\textbf{\bskblue{Methods}}
& \textbf{\bskblue{LLaMA-3.2-3B-Instruct}}
& \textbf{\bskblue{Qwen3-4B}} 
& \textbf{\bskblue{Qwen2.5-Instruct-3B}}  \\ 
\midrule
\bskblue{Init} &\bskblue{65.5}	&\bskblue{82.6}	&\bskblue{77.2}  \\
\bskblue{TTRL} &\bskblue{71.9} &\bskblue{87.5}	&\bskblue{78.9}  \\
\bskblue{TTVS} &\bskblue{73.3} &\bskblue{90.7}	&\bskblue{81.2} \\
\bottomrule
\end{tabular}}
\caption{\bskblue{Performance Analysis on OpenBookQA.}}
\vspace{-2mm}
\label{tab:openbook}
\end{table}

\bskblue{\subsection{Applicability to Open-Ended Tasks}\label{text:open}
To explore the boundaries of our method's applicability beyond purely mathematical reasoning, we conducted an additional experiment on OpenBookQA in Table \ref{tab:openbook}. This benchmark assesses a model's ability to reason by combining given facts with commonsense knowledge, representing a step towards more open-ended reasoning.
TTVS maintains a consistent and significant performance advantage over TTRL even in this commonsense reasoning context, suggesting that the benefits of our approach are not strictly limited to mathematical logic.}

\bskblue{\subsection{Discussion for "Non-verifiable" Scenarios} \label{text: Non-verifiable}
While the previous implementation of TTVS primarily focuses on domains with objective, verifiable rewards (e.g., mathematics), we acknowledge the challenge of extending this framework to open-ended tasks such as creative writing or summarization. In these "non-verifiable" scenarios, the absence of a rule-based or ground-truth reward signal complicates both the pseudo-labeling and the online filtering processes.
To bridge this gap, a promising direction involves transitioning from rule-based verifiers to model-based supervision, specifically through an LM-as-a-Judge paradigm. In this setup, a frozen, superior model (e.g., GPT-4o or a specialized Reward Model) could serve as the "verifiable signal" by evaluating the semantic consistency of synthesized query variants and the quality of reasoning rollouts. Specifically, the "Judge" can verify whether a synthesized creative prompt remains semantically equivalent to the original and use consensus-based scoring or pairwise comparisons to generate the necessary reward signals for self-exploration. Although this introduces considerations regarding the judge's bias and additional API costs, it provides a viable pathway to unlock the self-evolution capabilities of TTVS in broader, more subjective linguistic domains.
}
\bskblue{\subsection{Prompt Template for Variational Synthesis} \label{text: prompt}
The quality and semantic equivalence of the generated variants are primarily ensured by our carefully designed prompt, as detailed in Figure \ref{fig:prompt}. The prompt provides explicit and strict constraints to the model, including: \textbf{Preserve Semantic Equivalence} and \textbf{Maintain Mathematical Precision}: These instructions enforce that the core logic, numbers, and solvability of the problem remain unchanged.
\textbf{Vary Syntactic Structure and Lexical Expressions}: This instruction explicitly encourages the model to generate diverse paraphrases rather than superficial copies.
This structured prompting serves as the critical factor of defense, guiding the model to produce high-quality and diverse outputs, which is a key advantage of our approach.}
\label{sec:appendix}

\end{document}